\documentclass[lnbip]{svmultln}
\makeindex          
\usepackage{graphicx}
\usepackage{hyperref}
\usepackage{wrapfig}
\usepackage{colortbl}
\usepackage[USenglish,english,american]{babel}
\usepackage[english]{babel}
\usepackage{booktabs}
\usepackage[utf8]{inputenc}
\usepackage{soul}
\usepackage{times}
\usepackage{algorithm}
\usepackage[noend]{algpseudocode}
\makeatletter
\def\BState{\State\hskip-\ALG@thistlm}
\makeatother

\usepackage{multirow}
\usepackage{wrapfig,lipsum,booktabs}

\usepackage{amsfonts}
\usepackage{amsmath}
\usepackage{amssymb}
\usepackage{booktabs}
\usepackage{caption}
\usepackage{cite}
\usepackage{comment}
\usepackage{float}
\usepackage[utf8]{inputenc}
\usepackage{subfig}
\usepackage{adjustbox}
\usepackage{todonotes}
\usepackage{wrapfig}
\usepackage{rotating}

\def\mi#1{\mathit{#1}}

\def\post#1{\ensuremath{{#1}\kern-.05ex\bullet}\,}
\usepackage{xcolor}

\usepackage{tikz}
\usetikzlibrary{arrows,backgrounds,calc,decorations.pathmorphing,decorations.pathreplacing,fit,matrix,patterns,petri,positioning, shapes}

\newcommand{\multiset}{B}

\renewcommand*\abstractname{abstract}

\usepackage{diagbox}
\begin{document}
	\mainmatter              

\vspace{-0.4cm}
	\title{Conformance Checking Approximation using Subset Selection and Edit Distance}
	\titlerunning{Conformance Checking Approximation}  
	%
	\vspace{-0.4cm}
\author{Mohammadreza Fani Sani\inst{1} \and Sebastiaan J. van Zelst\inst{1,2} \and Wil M.P. van der Aalst\inst{1,2}}

\authorrunning{Mohammadreza Fani Sani et al.}   
%
\tocauthor{Mohammadreza Fani Sani, Sebastiaan J. van Zelst, Wil M.P. van der Aalst}
\institute{$^1$Process and Data Science Chair, 
	RWTH Aachen University, Aachen, Germany\\ 
	$^2$Fraunhofer FIT, Birlinghoven Castle, Sankt Augustin, Germany \\	
	\email{{fanisani, s.j.v.zelst, wvdaalst}@pads.rwth-aachen.de}}

 

	\maketitle          
	\vspace{-0.3cm}      	
	\begin{abstract}
Conformance checking techniques let us find out to what degree a process model and real execution data correspond to each other. 
In recent years, alignments have proven extremely useful in calculating conformance statistics.
Most techniques to compute alignments provide an exact solution.
However, in many applications, it is enough to have an approximation of the conformance value.
Specifically, for large event data, the computing time for alignments is considerably long using current techniques which makes them inapplicable in reality.
Also, it is no longer feasible to use standard hardware for complex processes. 
This paper proposes new approximation techniques to compute approximated conformance checking values close to exact solution values in a faster time. 
Those methods also provide upper and lower bounds for the approximated alignment value. 
Our experiments on real event data show that it is possible to improve the performance of conformance checking by using the proposed methods compared to using the state-of-the-art alignment approximation technique.
Results show that in most of the cases, we provide tight bounds, accurate approximated alignment values, and similar deviation statistics.
	
	\end{abstract}

	\section{Introduction}
	\label{sec:intro}

   One of the main branches of process mining is conformance checking, aiming at checking conformity of the discovered/designed process model w.r.t, real process executions~\cite{aalst_2016_pm}. 
   This branch of techniques is beneficial to detect deviations and to measure how accurate the discovered model is.
   In particular, the techniques in this branch are able to check conformance based on process modeling formalisms that allow for describing concurrency, i.e., the possibility to specify order-independent execution of activities.
   Early conformance checking techniques, e.g., “token-based replay”~\cite{rozinat_2008_conformance}, often lead to ambiguous and/or unpredictable results. 
   Hence, alignments~\cite{adriansyah_2012_alignment} were developed with the specific goal to explain and quantify deviations in a non-ambiguous manner. 
   Alignments have rapidly turned into the de facto standard conformance checking technique~\cite{Bas_2017_online}.
   Moreover, alignments serve as a basis for techniques that link event data to process models, e.g., they support performance analysis, decision mining~\cite{Leoni_2013_data}, business process model repair~\cite{fahland_2015_model} and prediction techniques.
   However, the computing alignments is time consuming on real large event data which makes it unusable in reality. 
   
   In many applications, we need to compute alignment values several times. 
   If we want to have a suitable process model for an event log we need to discover many process models using various process discovery algorithms with different settings, and measure how each process model fits with the event log using alignment techniques. 
   As normal alignment methods require considerable time for large event data, analyzing many candidate process models will be impractical.
   Consequently, by decreasing the alignment computation time, we can consider more candidate models in a limited time.
   
   In several cases, we do not need to have accurate alignment values, i.e., it is sufficient to have the approximated value or a close lower/upper bound for it. 
   Sometimes it is valuable to have a quick approximated conformance value and it is excellent worth to let users adjust the level of approximation.
   
   In this paper, we propose several conformance checking approximation methods that provide approximated alignment values plus lower and upper bounds for the actual alignment value. 
   These methods are capable to return problematic activities based on their deviation rates. 
   They work based on considering just a subset of the process model behavior instead of its all its behavior. 
   Using these methods, users are able to adjust the amount of process model behaviors considered in the approximation which affects the computation time and the accuracy of alignment values and their bounds.
   
   We implemented collections of methods in two open-source process mining tools and applied them on several large real event data and compared them with the state-of-the-art alignment approximation method.
   The results show that using some of proposed methods, we are able to approximate alignment values faster and at the same time the approximated values are very close to actual alignment values.

   The remainder of this paper is structured as follows. 
   In \autoref{sec:related_work}, we discuss related work.
   \autoref{sec:preliminaries} defines preliminary notation that eases the understanding of the proposed method that is explained in \autoref{sec:Method}.
   The evaluation and its results are given in \autoref{sec:eval}.
   Finally, \autoref{sec:conclusion} concludes the paper and presents some future work.

	\section{Related Work}
	\label{sec:related_work}
	
A plethora of different process mining techniques exists, ranging from process discovery to prediction. However, given the focus of this paper, we limit related work to the field of conformance checking computation and sampling techniques in the process mining domain. We refer to~\cite{aalst_2016_pm} for an overview of different process mining techniques.

In \cite{elhagaly_2019_evolution}, the authors review the conformance checking techniques in process mining domain.
\cite{carmona_2018_conformance} studies different methods for conformance checking, its applications and the software that provide it. 
Early work in conformance checking uses token-based replay~\cite{rozinat_2008_conformance}. 
The techniques replay a trace of executed events in a process model (Petri net) and add missing tokens if transitions are not able to fire. 
After replay, remaining tokens are counted and a conformance statistic is computed based on missing and remaining tokens. Alignments were introduced in~\cite{van_2012_replaying} and have rapidly developed into the de facto standard for conformance checking. 
In~\cite{aalst_2013_decomposing  ,munoz_2014_single}, decomposition techniques are proposed together with computing alignments.
Moreover, \cite{lee_2018_recomposing} proposed the decomposition method to find an approximation of the alignment in a faster time.
Applying decomposition techniques improves computation time, i.e. the techniques successfully use the divide-and-conquer paradigm; however, these techniques are beneficial when there are too many unique activities in the process~\cite{verbeek_2017_divide}.
More recently, general approximation schemes for alignments, i.e. computation of near-optimal alignments, have been proposed in~\cite{taymouri_2016_recursive}.
Finally, the authors in \cite{Bas_2017_online} proposed to incrementally compute prefix-alignments, paving the way for real-time conformance checking for event data streams.
Their approach can provide both optimal and approximate alignment values.

A relatively limited amount of work has been done to use sampling approaches in process mining domain. 
In \cite{carmona_2010_parikh}, the authors proposed a sampling approach based on Parikh vectors of traces to detect the behavior in the event log.
In~\cite {bauer_2018_much}, the authors recommend a trace-based sampling method to decrease the discovery time and memory footprint that assumes the process instances have different behavior if they have different sets of directly follows relations. 
Furthermore, \cite{berti_2017_sampling} recommends a trace-based sampling method specifically for the Heuristic miner\cite{weijters_2011_fhm}.
In both of these sampling methods, we have no control on the size of the final sampled event data.
Also, depend on the defined behavioral abstraction,the methods may select almost all the process instances.
Finally, all these sampling methods are unbiased and consequently they leads to non-deterministic results. 
In \cite{sani_2018_repairing}, we analyze random and biased sampling methods with which we are able to adjust the size of the sampled data for process discovery.

Some research has been done to approximate the alignment value.
\cite{bauere_2019_stimating} proposes sampling the event log and applying the conformance checking algorithm on the sampled data. 
The method increases the sample size until the approximated value is accurate enough. 
But, the proposed method does not guarantee the accuracy of the approximation, for example by providing bounds for it. 
In \autoref{sec:eval}, we show that if there is lot of unique behavior in the event log, using this method, the approximation time exceeds the computation time for finding the alignment value.
The authors in \cite{padro_2019_approximate} proposed a conformance approximation method that applies relaxation labeling methods on a partial order representation of a process model. 
Similar to the previous method, it does not provide any guarantee for the approximated value.
Furthermore, it needs to preprocess the process model each time.  
In this paper, we propose multiple alignment approximation methods that increase the conformance checking performance.
The methods also provide upper and lower bounds for alignment values.
Using the proposed methods the user is able to adjust the performance and the accuracy of the approximation.

\vspace{-0.3cm}
	
	\section{Preliminaries}
	\label{sec:preliminaries}

In this section, we briefly introduce basic process mining and, specifically, conformance checking terminology and notations that ease the readability of this paper\footnote{For some concepts, e.g., labeled Petri net and alignment please use the definitions in~\cite{aalst_2013_decomposing}}.

\begin{figure}[tb]
	\centering
	\begin{minipage}{.45\textwidth}
		\centering
		\includegraphics[width=1\textwidth]{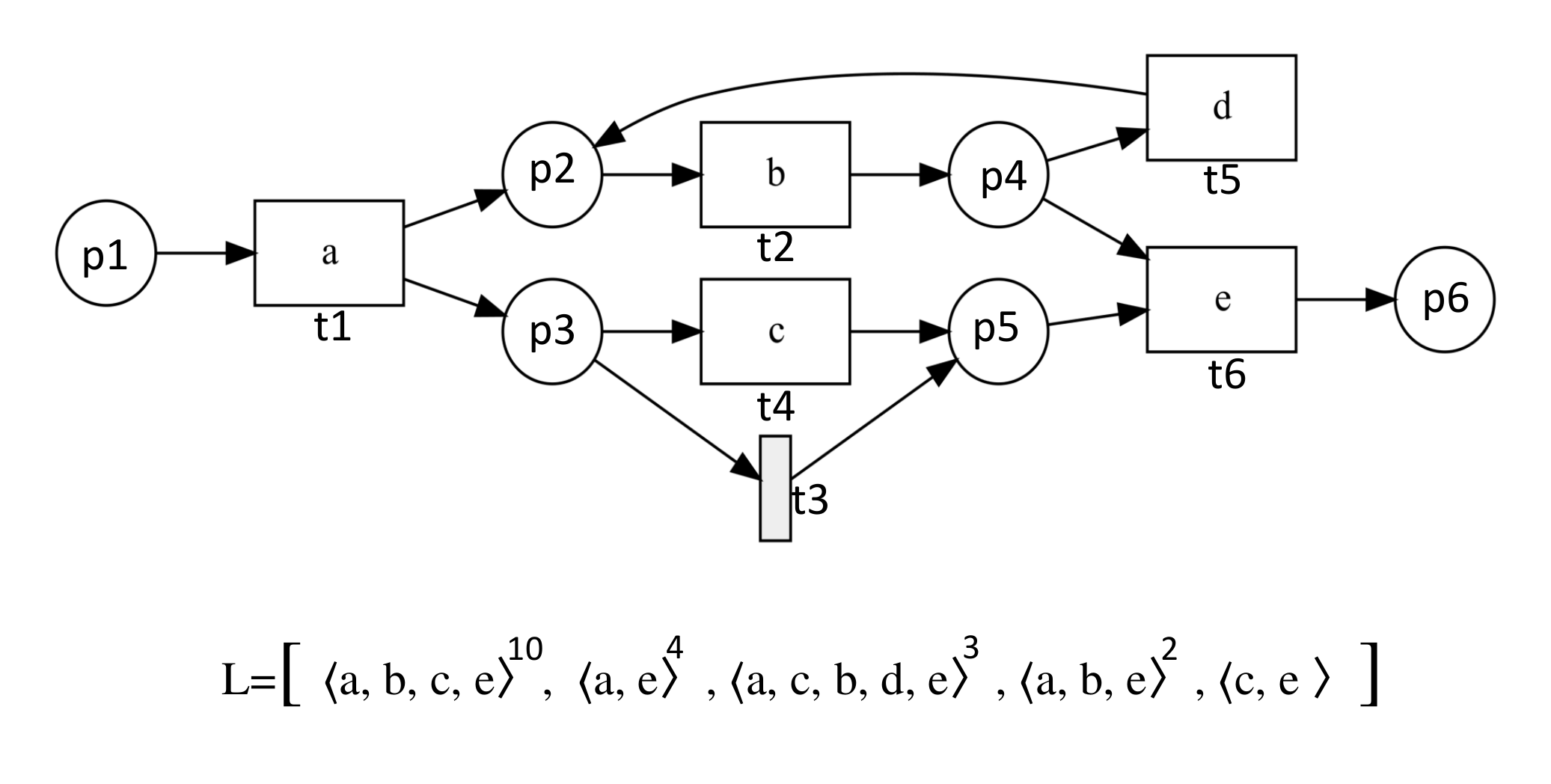}
		\caption{An example Petri net and an event log in a multiset view.
			\label{fig:example}
		}
	\end{minipage}%
	\begin{minipage}{.48\textwidth}
		\centering
		\includegraphics[width=1\textwidth ]{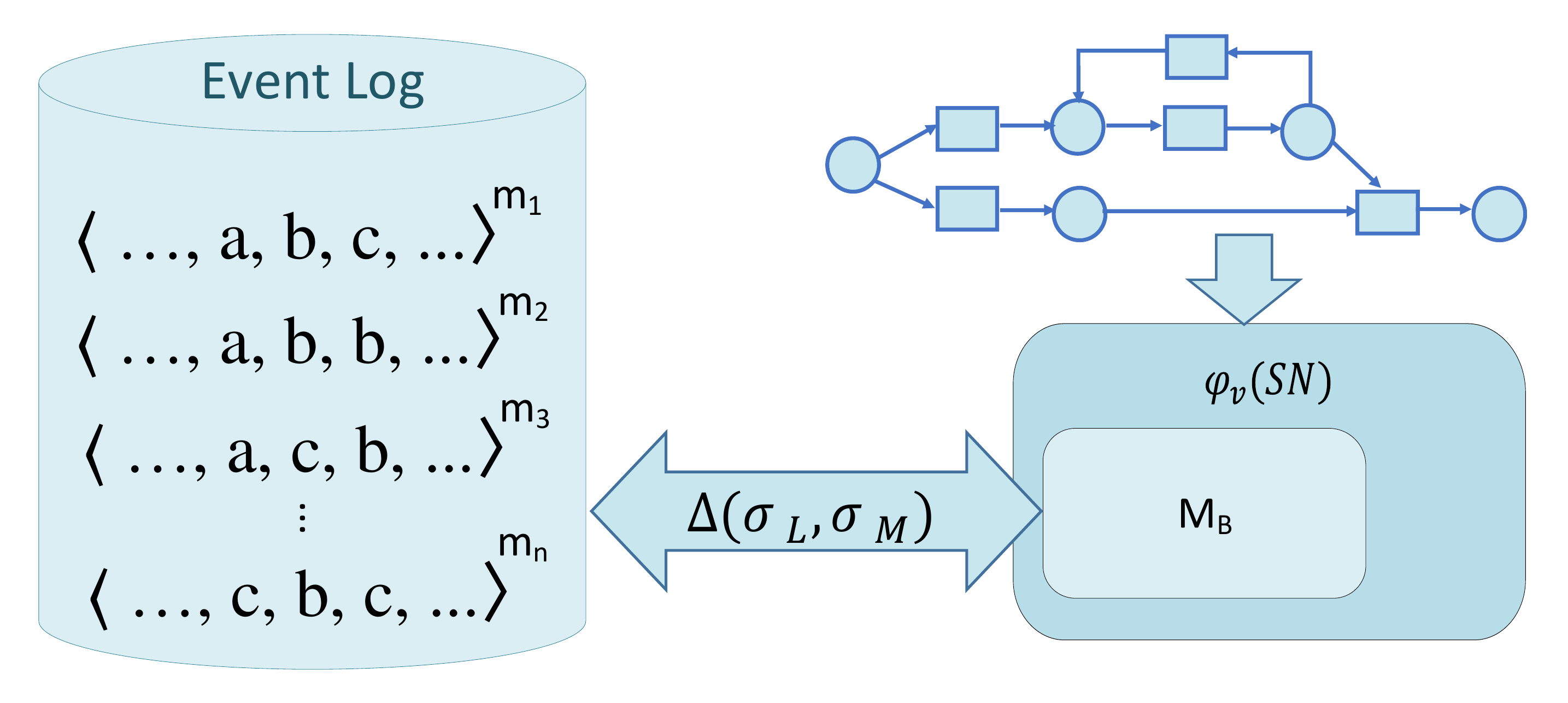}
		\caption{Overview of the proposed approximation approach. It uses $ M_B \subseteq \phi_v(SN) $ to approximate the alignment cost. 
			\label{fig:ProposedMethod} }
	\end{minipage}
\vspace{-0.4cm}
\end{figure}

Given a set $X$, a multiset $\multiset$ over $X$ is a function $\multiset \colon X \to \mathbb{N}_{\geq0}$, i.e., it allows certain elements of $X$ to appear multiple times.
We write a multiset as $\multiset = [e^{k_1}_1, e^{k_2}_2, ..., e^{k_n}_n]$, where for $1 \leq i \leq n$ we have $\multiset(e_i) = k_i$ with $k_i \in \mathbb{N}_{>0}$. 
If $k_i = 1$, we omit its superscript, and if for some $e \in X$ we have $\multiset(e) = 0$, we omit it from the multiset notation.
$\overline{\multiset} = \{ e \in X \mid \multiset(e) > 0 \}$ is the set of elements present in the multiset.
The set of all possible multisets over a set $X$ is written as $\mathcal{\multiset}(X)$. 
Some examples: $\multiset_1 = [~]$, $\multiset_2 = [a^2,b]$, $\multiset_3 = [a,c]$, $\multiset_4 = [a^3,b,c]$
are multisets over $X=\{a,b,c\}$. 
The standard set operators can be extended to multisets, e.g., $a \in \multiset_2$, $\multiset_2 \uplus \multiset_3 = \multiset_4$, $\multiset_5 \setminus \multiset_2 = \multiset_3$, $|\multiset_4|=5$, etc.
$\{x\in \multiset\}$ denotes the set of all elements $x\in X$ for which $\multiset(x) \geq 1$.
$[f(a) \mid a \in \multiset]$ denotes the multiset where element $f(a)$ appears $\sum_{x \in \multiset \mid f(x)=f(a)}\multiset(x)$ times.
For example, if $ f(a)=f(c)\neq f(b) $, $[f(x) \mid x \in \multiset_4]=[f(a)^4, f(b)] $.


Given a system net $ SN $, $\phi_f(\mi{SN})$ is the set of all complete firing sequences of $ SN $ and  $\phi_v(\mi{SN})$ is the set of all possible \emph{visible} traces,
i.e., complete firing sequences starting its initial marking and ending in its final marking projected onto the set of observable activities (not silent transitions e.g., $ t_3 $ in \autoref{fig:example}).

To measure how a trace aligns to a process model, we need to define the notation of moves.
A \emph{move} is a pair $(x,t)$ where the first element refers to the log and the second element refers to the corresponding transition in the model.
For example, $(a,t_1)$ means that both log and model make an ``$a$ move''
and the move in the model is caused by the occurrence of transition $t_1$ (the label of $t_1$ is \textit{a}).

An alignment is a sequence of legal moves such that after removing all $\gg$ symbols,
the top row corresponds to a trace in the event log and the bottom row corresponds
to a complete firing sequence of in $ \phi_f(SN) $.
The middle row corresponds to a visible path when ignoring the $\tau$ steps, i.e., corresponding to silent transitions (e.g., $ t_3 $ in \autoref{fig:example}).
For silent transitions, there is no corresponding recorded event in the log.

\begin{definition}[Alignment]
	Let $ \sigma_L \in L$ be a log trace and $ \sigma_M \in \phi_f(SN) $ a complete firing sequence of a system net $ SN $. 
	An alignment of $ \sigma_L  $ and $ \sigma_M $ is a sequence of pairs $ \gamma \in A_{LM}^* $ such that the projection on the first element (ignoring $ \gg $) yields $ \sigma_L $ and the projection on the second element (ignoring $ \gg $ and transition labels) yields $ \sigma_M $.
\end{definition}	

The following alignments relate to $ \sigma_L= \langle a,c,b,d,e \rangle $ and the model in \autoref{fig:example}.
\scriptsize
$$
\gamma_1 = \begin{array}{|c|c|c|c|c|c|}
a &\gg & c & b & d & e \\ \hline
a & \tau &\gg & b & \gg & e  \\
t_1 & t_3 & & t_2 &  & t_6  \\
\end{array}
\ \ 
\gamma_2 = \begin{array}{|c|c|c|c|c|c|c|}
a & c & b & d &  e \\ \hline
a & c & b & \gg  &  e  \\
t_1 & t_4 & t_2 & &  t_6  \\
\end{array}
$$
\normalsize

By considering the label of visible transitions of alignments, we are able to find the corresponding model trace, e.g., the model trace of $ \gamma_1 $ is  $\langle a,b,e \rangle $.
Given a log trace and a process model, it is possible to have many alignments.
To select the most appropriate one, we associate
\emph{costs} to different move types and select an alignment with the lowest total costs.
To quantify the costs of misalignments we introduce a cost function $\delta$.
\begin{definition}[Cost of Alignment]	
	Cost function $ \delta \in A_{LM} \to \mathbb{R}\geq 0$ assigns costs to legal moves.
	The cost of an alignment $ \gamma \in A_{LM}^* $ is $ \delta(\gamma)= \Sigma_{(x,y)\in \gamma} \delta(x,y) $.
\end{definition}
Synchronous moves that are similar in the trace and the model have no costs.
In other words, for all $x\in \mathcal{U}_A$, $\delta(x,t) = 0 $ if $l(t)=x  $, are called synchronous moves.
Moves in model only have no costs if the transition is invisible, i.e., $\delta(\gg,t)=0$ if $l(t) = \tau$.
Moreover, $\delta(\gg,t)>0$ is the cost when the model makes an ``$x$ move''
In this paper we use a standard cost function $\delta_S$
that assigns unit costs:
$\delta_S(\gg,t) = \delta_S(x,\gg) = 1$ if $ l(t) \neq \tau $.
In the above example alignments, $ \delta_S(\gamma_1)=2 $ and $ \delta_S(\gamma_2)=\delta_S(\gamma_3)=1 $.

\begin{definition}[Optimal Alignment]
	Let $ L\in \multiset(\mathcal{U}_A^*) $ be an event log and let $ SN$ be a system net with $ \phi_v(SN)\neq\emptyset $.
	
	- For $ \sigma_L \in L $, $ \Gamma_{\sigma_L, SN} = \{ \gamma\in A_{LM}^* | \exists_{\sigma_M \in \phi_f(SN)}$ is an alignment of $\sigma_L$ and $\sigma_M  \}$.  
	
	- An alignment $ \gamma \in \Gamma_{\sigma_L, SN}  $ is optimal for trace $ \sigma_L \in L $ and system net $ SN $ if for any alignment $ \gamma' \in \Gamma_{\sigma_L, M}: \delta(\gamma') \geq \delta ( \gamma)$.
	
	- $ \gamma_{SN} \in \mathcal{U}_A^* \to A_{LM}^* $ is a mapping that assigns any log trace $ \sigma_L $ to an optimal alignment, i.e., $ \gamma_{SN}(\sigma_L) \in \Gamma_{\sigma_L, SN}  $ and $ \gamma_{SN}(\sigma_L) $ is an optimal alignment.
	
	- $ \lambda_{SN} \in \mathcal{U}_A^* \to \mathcal{U}_A^* $ is a mapping that assigns any log trace $ \sigma_L $ to visible activities of the model trace of the optimal alignment.
\end{definition}		
In the running example, $\gamma_{SN}(\langle a,c,b,d,e \rangle)$ equals to $ \gamma_2 $ is an optimal alignment, and $ \lambda(\sigma_L) = \langle a,c,b,e\rangle $ is the corresponding model trace for the optimal alignment. 

We are able to compute the distance of two traces (or two sequences) in a faster way using the adapted version of Levenshtein edit distance~\cite{sellers_1974_theory}.
Suppose that $ \sigma,\sigma' \in A^* $, Edit Distance function $ \bigtriangleup (\sigma, \sigma')\to \mathbb{N}$ is the minimum number of edits that are needed to transform $ \sigma  $ to $ \sigma' $. 
As edit operations, we allow a deletion or an insertion of an activity (or a transition label) in a trace.
To give an example, $ \bigtriangleup(\langle a,c,f,e,d \rangle,$ $ \langle a,f,c,a,d \rangle  )=4 $ corresponding to two deletions and two insertions.
Note that this measure is symmetric, i.e., $ \bigtriangleup (\sigma, \sigma')= \bigtriangleup (\sigma', \sigma) $.

We can show that it is possible to use the edit distance function instead of the standard cost function. 
In other words, $ \bigtriangleup $ and $ \delta_S $ return same distance values.
We are able to extend the $ \bigtriangleup $ function from unit cost (i.e., $ \delta_S $) to any other cost by giving different weights to insertion and deletion of different activities.

In~\cite{marzal1_993_computation} it is explained that the Levenshtein metric before normalization satisfies the triangle inequality. 
In other words, $ \bigtriangleup (\sigma , \sigma' ) \leq \bigtriangleup (\sigma , \sigma'' ) + \bigtriangleup (\sigma'' , \sigma' ) $. 
Moreover, suppose that S is a set of sequences, $ \Phi (\sigma_L, S)= \min\limits_{\sigma_M \in S}\bigtriangleup(\sigma_L, \sigma_M) $ returns the distance of the most similar sequence in $ S $.

Let $ \phi_v(SN) $ is a set of all visible firing sequences in $ SN $, and $ \gamma_{SN} (\sigma) $ is an optimal alignment for sequence $ \sigma $. 
It is possible to prove that $ \delta_S(\gamma_{SN}(\sigma))= \Phi(\sigma,\phi_v (SN) ) $\footnote{Because of the page limit, we do not provide the proof here. }.

Using the edit distance, we are also able to find which activities are required to be deleted or inserted. 
So, not only the cost of alignment; but, the deviated parts of the process model (except invisible transitions) are also detectable using this function.

It is possible to convert misalignment costs into a fitness value using \autoref{equ:fitness}.
It normalizes the cost of optimal alignment by one deletion for each activity in the trace and one insertion for each visible transition in the shortest path of model (SPM). 
The fitness between an event log and a system net is a wighted average of traces' fitness. 
\vspace{-0.3cm}
\begin{equation}
\scriptsize
\label{equ:fitness}
fitness(\sigma_L,SN)= 1-  \frac{\delta(\gamma_{SN}(\sigma))}{|\sigma_L| + \min\limits_{\sigma_M \in \phi_f} (|\sigma_M|)}
\end{equation}

\section{Approximating Alignments using Subset of Model Behavior}
\label{sec:Method}
As computational complexity of computing alignment is exponential in the number of states and the number of transitions, it is impractical for larger petri nets and event logs~\cite{adriansyah_2015_measuring}.
Considering that the most time consuming part in conformance checking procedure is finding the optimal alignment for each $ \sigma_L \in L $ and the system net $ SN $ leads us to propose an approximation approach that requires fewer alignment computations.
The overview of the proposed approach is presented in \autoref{fig:ProposedMethod}.
We suggest to use $ M_B \subseteq \phi_v (SN) $ instead of the whole $ \phi_v(SN) $ and apply the edit distance function instead of $ \delta_S $.  
In the following lemma, we show that using this approach, we have an upper bound for the cost of alignment (i.e., a lower bound for fitness).

\begin{lemma}[Alignment Cost Upper Bound]
	Let $ \sigma_L \in \mathcal{U}_A^*$ is a log trace and, and $ \sigma_M \in \phi_v (SN) $ is a visible firing sequence of SN.
	We have $ \delta_S (\gamma_{SN}(\sigma_L))\leq \bigtriangleup (\sigma_L, \sigma_M) $ where $ \gamma_{SN}(\sigma_L) $ is the optimal alignment.

\end{lemma}
\textbf{Proof:}
We shown that $ \bigtriangleup(\sigma_L, \sigma_M)= \delta_S(\gamma) $, so we have $ \bigtriangleup(\sigma_L, \sigma_M) \geq \delta_S(\gamma_{SN}(\sigma_L)) $.
Therefore, if $ \delta_S (\gamma_{SN}(\sigma_L))> \bigtriangleup (\sigma_L, \sigma_M) $, $  \gamma_{SN}(\sigma_L) $ is not an optimal alignment. 
Consequently, if we use any $ M_B\subseteq \phi_v(SN) $, $ \Phi (\sigma_L, M_B) $ returns an upper bound for the cost of optimal alignment.

Here, we explain two main components of our proposed approach, i.e., constructing a subset of model behavior ($ M_B $) and computing the approximation.

\subsection{Constructing Model Behavior }
As explained, we propose to use $ M_B $ i.e., a subset of visible model traces to have an approximated alignment.
An important question is how to construct  $ M_B $.
In this regard, we propose two approaches, i.e., \textit{simulation} and \textit{candidate selection}.

\textbf{1) Simulation:} The subset of model traces can be constructed by simulating the process model.
In this regard, having a system net and the initial and final markings, we simulate some complete firing sequences. 
Note that, we keep only the visible firing sequences in $ M_B $.
It is possible to replay the Petri net randomly or by using more advanced methods, e.g., stochastic petri net simulation techniques. 
This approach is fast; but, we are not able to guarantee that by increasing the size of $ M_B $ we will obtain the perfect conformance (or fitness) value, because the model traces are able to be infinite.  
Another potential problem of this method is the generated subset may be far from traces in the event log that leads to have inaccurate approximation.  

\textbf{2) Candidate Selection:} The second method to construct $ M_B $ is computing the optimal alignments of selected traces in the event log and finding the corresponding model traces for these alignments. 
In this regard, we first select some traces (i.e., candidates) from the event log $L $ and put them in $ L_C $.
Then for each $ \sigma_L \in L_C $ we find the optimal alignment and insert $ \lambda_{SN} (\sigma_L) $ to $ M_B $.
Thereafter, for other traces $\sigma_L'\in L_C' $ (i.e., $ L_C'= L- L_C $), we will use $ M_B $ and compute $ \Phi (\sigma_L', M_B ) $.

As the edit distance function holds triangle inequation property, it is better to insert $ \lambda_{SN} (\sigma_) $ in $ M_B $ instead of considering $ \sigma_L $.
To make it more clear, let $ \sigma_L $ is a log trace, $ SN $ is a system net, and $\sigma_M= \lambda_{SN} (\sigma_L) $ is the corresponding visible model trace for an optimal alignment of $  \sigma_L $ and $ SN $.
According to triangle inequation property, for any trace $ \sigma \in L$, we have $ \bigtriangleup (\sigma, \sigma_M) \leq \bigtriangleup(\sigma, \sigma_L ) + \bigtriangleup(\sigma_L, \sigma_M ) $.	
So, the cost of transforming $ \sigma_L $ to $ \sigma_M $ is less than the cost of transforming it to $ \sigma_L $ and then to  $ \sigma_M $.
As $ \Phi (\sigma_L, M_B)  $ returns the minimum cost of the most similar sequence in $ M_B $ to $ \sigma_L $, putting directly the alignments of traces $ M_B $ causes to have a smaller upper bound for alignment cost. 
Moreover, it is possible to have $  \lambda_{SN}(\gamma_{SN}(\sigma_1)) =\lambda_{SN}(\gamma_{SN}(\sigma_2)) $ for $ \sigma_1 \neq \sigma_2 $.
Therefore, by inserting $  \lambda_{SN}(\sigma_1)$ instead of $ \sigma_1 $ in $ M_B $, we will have $ M_B $ with fewer members that increases the performance of the approximation.

To select the candidate traces in $ L_C $, we propose three different methods.
We can select these traces \textit{randomly} or based on their \textit{frequency} in the event log (i.e, $ L(\sigma_L) $).
The third possible method is to apply a \textit{clustering} algorithm on the event log and put the traces in $ K $ different clusters based on their control follow information. 
We then select one trace,i.e., medoid for each cluster representing all members. 
It is expected that by using this approach, the detected bounds will be more accurate.

\vspace{-0.1cm}
\subsection{Computing Alignment Approximation }
After constructing $ M_B $, we use it for all traces in the $ L_C' $.
Note that for the \textit{simulation} method, $ L=\emptyset$ and $ L_C'=L $. 
Moreover, for the \textit{candidate selection} method, we use the alignment values that already computed by in constructing $ M_B $.
To compute the lower bound for the fitness value, we compute the fitness value of all of the $\sigma \in L_C' $ using $ \Phi(\sigma,M_B) $.
Afterwards, based on the weighted average of this fitness and alignments that are computed in the previous part, the lower bound for the fitness value is computed. 

For the upper bound of fitness value, we compare the length of each trace in $L_C'  $ with the shortest path in the model (i.e., $ SPM $).
To find $ SPM $, we compute the cost of optimal alignment for an empty trace (i.e., $ \epsilon $) and the system net.
In the example that is given in \autoref{fig:example}, $ SPM=3 $.
If the length of a trace is shorter than the \textit{SPM}, we know that it needs at least $ SPM-\sigma_L $ edit operations (i.e., insertions) to transform to one of model traces in $ \phi_f(SN) $. 
Otherwise, we consider at least 0 edit operation for that trace.
Because, it is possible that there is a model trace that $ \sigma_M \in \phi_v(SN)  $ and $ \sigma_M \notin M_B $ and it perfectly fits to the log trace.  
After computing the upper bound values for all traces in $ L_C' $, based on the weighted average of them and the computed fitness values of traces in $ L_C $ we compute the upper bound value for fitness.

\begin{table*}[tb]
	\center
	\caption{Result of using the proposed approximation method for the event log that is given in \autoref{fig:example} considering that $ M_B= \{ \langle a,b,e \rangle, \langle a,b,c,e \rangle  \} $. }
	\label{tab:exam}
	\scriptsize
	\begin{tabular}{|l|c|c|c|c|c|c|c|}
		
		\hline
		Trace	& $ \delta_S(\gamma_{SN}) $ & Min $\bigtriangleup $ & Actual Fitness & LBoundFitness & UBoundFitness & AppxFitness& Freq \\ \hline
		$\langle a,b,c,e \rangle$ & 0 & 0 & 1 & 1 & 1 & 1& 10  \\ \hline
		$\langle a,e \rangle$ & 1 & 1 & 0.8 & 0.8 & 0.8 & 0.8 & 4 \\ \hline
		$\langle a,c,b,d,e \rangle$ & 1 & 2 & 0.875 & 0.75 & 1 & 0.875 & 3 \\ \hline
		$\langle a,b,e \rangle$ & 0 & 0 & 1 & 1 & 1 & 1 &  2 \\ \hline
		$\langle c, e \rangle$ & 2 & 2 & 0.5 & 0.5 & 0.8 & 0.65 & 1  \\ \hline
		$ L  $ & $\sim$ & $\sim$ & 0.916 & 0.898 & 0.95 & 0.924 & $\sim$ \\ \hline
	\end{tabular}
\vspace{-0.1cm}
\end{table*}

To compute the approximation values, for each trace in $\sigma \in L_C' $, we compute the $ \Phi(\sigma, M_B) $ and compare it to the average fitness value of $L_C  $. 
If the fitness value of the new trace is higher than $ Fitness(L_C,SN) $, we consider $ \Phi(\sigma, M_B) $ as the approximated fitness value; otherwise, $ Fitness(L_C,SN) $ will be considered for the approximation.
Similar to the bounds, we use the weighted averages of fitness values of $ L_C $ and $ L_C' $ to compute the approximated fitness value of whole event log.
Note that, for the simulation method that $ L_C=\emptyset $, the approximated fitness value for each trace (and for the whole event log) is equal to the lower bound.

Finally, the proposed method returns the number of asynchronous (i.e., deletions and insertions) and synchronous moves for each activity in the event log. 
This information helps the data analyst to find out the source of deviations.

The computed bounds and the approximated fitness value for each trace and the overall event log in \autoref{fig:example} based on $ M_B= \{ \langle a,b,e \rangle, \langle a,b,c,e \rangle  \} $ is given in \autoref{tab:exam}.
This $ M_B $ is possible to gained by computing the alignment of the two most frequent traces in the event log or by simulation.
The approximated fitness will be $ 0.924 $ that its accuracy equals to $ 0.008 $. 
The proposed bounds are $  0.924$ and $ 0.898 $. 
Moreover, the method returns the number of insertion and deletions that are 1 insertion for \textit{a}, 5 insertion for \textit{b}, 3 deletions for \textit{c}, 3 deletions for \textit{d}, and nothing for \textit{e}.

By increasing $ |M_B| $ we have more accurate approximations and bounds.
However, increasing the $ |M_B| $ for candidate selection approach increases the number of required alignments computations and consequently increasing the computation time.

\vspace{-0.1cm}
\section{Evaluation}
\label{sec:eval}
\vspace{-0.1cm}
In this section, we aim to explore the accuracy and the performance of our methods. 
We first explain the implementation of them and afterward, we explain the experimental setting.
Finally, the experimental results and some discussions will be provided. 
\vspace{-0.1cm}
\subsection{Implementation}
To apply the proposed conformance approximation method, we implemented the \textit{Conformance Approximation} plug-in in the \texttt{ProM}~\cite{van_2009_prom} framework\footnote{\small \scriptsize\url{svn.win.tue.nl/repos/prom/Packages/LogFiltering} }.
It takes an event log and a Petri net as inputs and returns the conformance approximation, its bound, and the deviation rates of different activities.
In this implementation, we let the user adjusts the size of $ M_B$ and the method to select and insert model traces in it (i.e., \textit{simulation} and alignment of selected candidates).
If the user decides to use alignments for creating model behavior, she can select candidates based on their \textit{frequency}, \textit{random}, or using the \textit{clustering} algorithm.
For finding the distance of a log trace and a model trace, we used the \textit{edit distance} function that is an adapted version of Levenshtein distance~\cite{sellers_1974_theory}.
To cluster traces, we implement the K-Medoids algorithm that returns one trace as a candidate for each cluster~\cite{de_2013_effective} based on their edit distance.

To apply proposed methods on various event logs and parameters, we ported the developed plug-in to \texttt{RapidProM} that is an extension of \texttt{RapidMiner} that combines scientific work-flows with a several process mining algorithms~\cite{Rad}.
\begin{table*}[tb]
	
	\centering
	\caption{Statistics regarding the real event logs that are used in the experiment.}
	\label{tab: EventLOgs}
	\scriptsize
	\vspace{-0.2cm}
	\begin{tabular}{|l|c|c|c|c|c|}
		\hline
		\textbf{Event Log} & \textbf{Activities\#} & \textbf{Traces\#} & \textbf{Variants\#} & \textbf{DF\#} & \textbf{Uniqueness} \\ \hline
		\textit{$ \text{BPIC-}2012$}~\cite{BPI_2012_challeng}   & 23 & 13087 & 4336 & 138 & 0.33\\ \hline
		\textit{$ \text{BPIC-}2018\text{-Department}$}~\cite{BPI_2018_challeng} & 6 & 29297 & 349 & 19 & 0.01 \\ \hline
		\textit{$ \text{BPIC-}2018\text{-Inspection}$}~\cite{BPI_2018_challeng}  & 15 & 5485 & 3190 & 67 & 0.58   \\ \hline
		\textit{$ \text{BPIC-}2018\text{-Reference}$}~\cite{BPI_2018_challeng}  & 6 & 43802 & 515 & 15 & 0.01  \\ \hline
		\textit{$ \text{BPIC-}2019 $}~\cite{BPI_2019_challeng}   & 42 & 251734 & 11973 & 498 & 0.05  \\ \hline
		\textit{$\text{Hospital-Billing} $}~\cite{mannhardt_2017_hospital}  & 18 & 100000 & 1020 & 143 & 0.01  \\ \hline
		\textit{$ \text{Road} $}~\cite{de_2015_road}   & 11 & 150370 & 231 & 70 & $ \sim  0 $ \\ \hline
		\textit{$ Sepsis $}~\cite{Sepcis_2016_Felix}  & 16 & 1050 & 846 & 115 & 0.81 \\ \hline
	\end{tabular}
	\vspace{-0.4cm}
\end{table*}

\vspace{-0.1cm}
\subsection{Experimental Setup}
\label{subsec:ES}
We applied the proposed methods on eight different real event logs. 
Some information about these event logs is given in \autoref{tab: EventLOgs}. Here, \textit{uniqueness} refers to $ \frac{Variant\#}{Trace\#} $.

For process discovery, we used the Inductive Miner~\cite{leemans_2014_ind_infreq} with infrequent thresholds equal $ 0.3 $, $ 0.5 $, and $ 0.7 $.
We applied conformance approximation methods with different settings.  
In this regard, an approximation parameter is used with values equal to $ 1 $, $ 2 $, $ 3 $, $ 5 $, $ 10 $, $ 15 $, $ 20 $, $ 25 $, and $ 30 $. 
This value for the \textit{Simulation} method is the number of simulated traces times $ |L|$, and for the  \textit{candidate selection} methods (i.e., \textit{clustering}, \textit{frequency}, and \textit{random}), it shows the relative number of selected candidates, i.e., $ \frac{|L_C|}{|L|} $. 
We also compared our proposed method with the \textit{statistical} sampling method~\cite{bauere_2019_stimating}.
The approximation parameter for this method determines the size and the accuracy of sampling.
We consider $ \text{epsilon} = \text{delta}= \text{approximation parameter} \times 0.001 $. 
We did not consider \cite{lee_2018_recomposing} in the experiments, as it does not able to improve the performance of normal computation of alignment~\cite{van_2018_efficiently} for event logs which have few unique activities using the default setting. 
Even for some event logs with lots of unique activities in \cite{lee_2018_recomposing}, the performance improvement of our methods is higher. 
Because of the page limit, we do not show results of this experiment here.

For process discovery, we used the Inductive miner algorithm~\cite{leemans_2014_ind_infreq} with the noise threshold equal to $ 0.3 $, $ 0.5 $, and $ 0.7 $.
In all experiments and for all methods, we used eight threads of CPU.   
Moreover, each experiment was repeated four times, since the conformance checking time is not deterministic, and the average values are shown.

\begin{figure*}[tb]
	
	\begin{center}
		
		\includegraphics[width=0.97\textwidth,height=4.8cm] {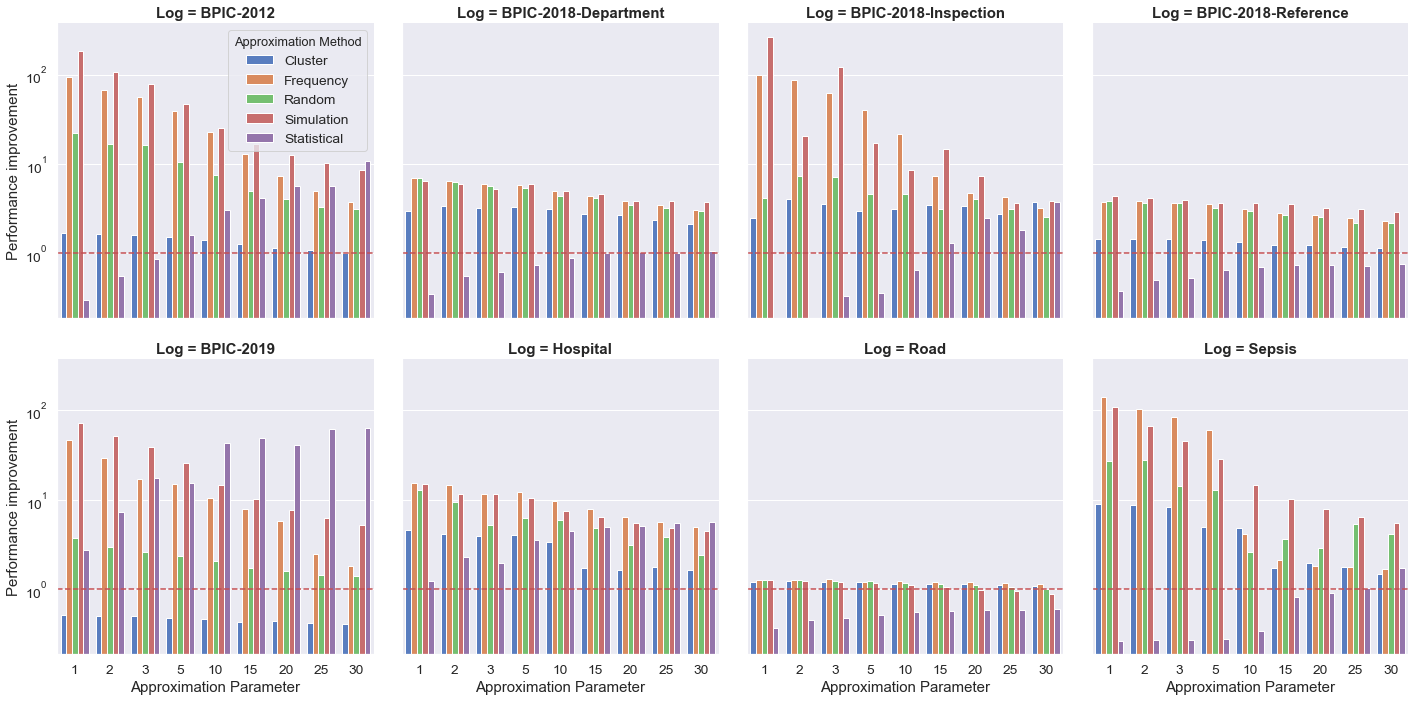}
		\caption{ Performance improvement with consideration of preprocessing time. 
			\label{fig:PerfImprove}
		}
	
		\includegraphics[width=0.97\textwidth,height=4.8cm] {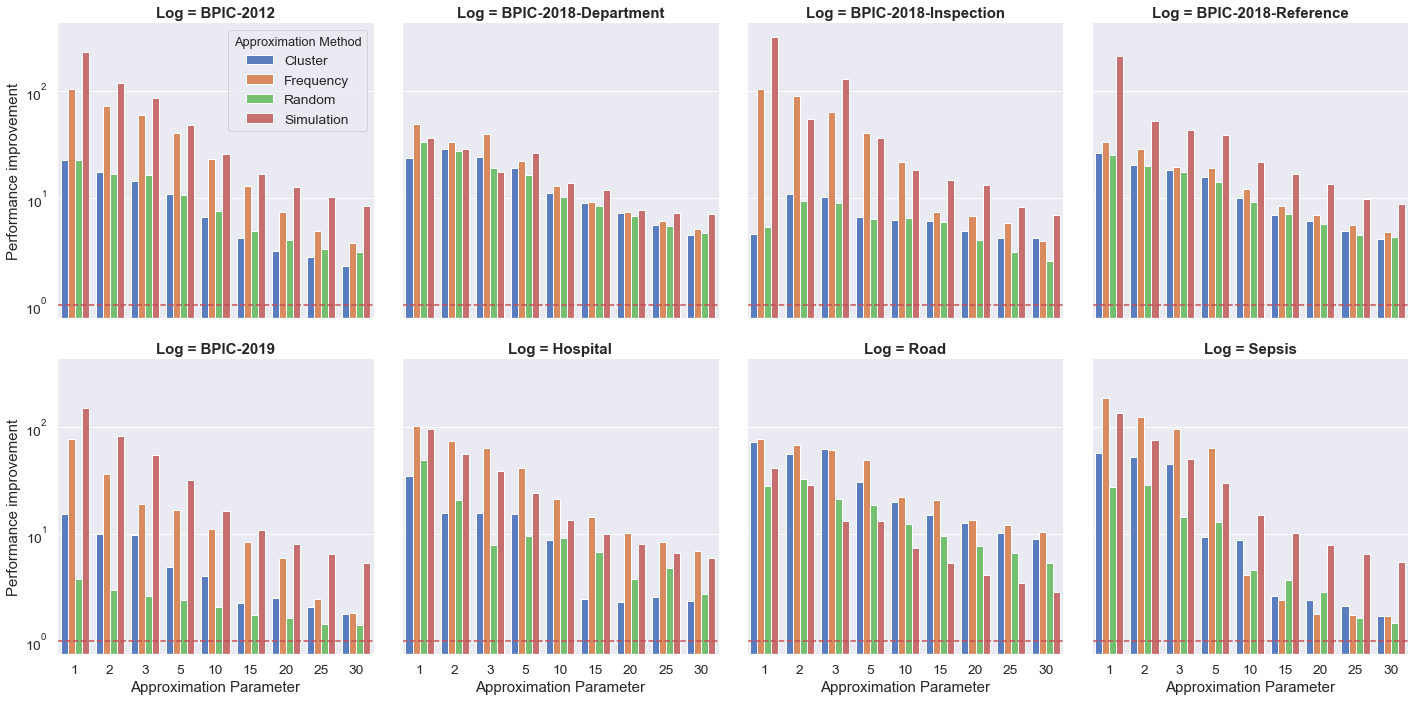}
		\caption{ Performance improvement without consideration of preprocessing time.  
			\label{fig:PerfImprove2}
		}
	\end{center}
	\vspace{-0.5cm}
\end{figure*}

To evaluate how the conformance approximation is able to improve the performance of the conformance checking process, we used the $ PI= \frac{\text{Normal Conformance Time}}{\text{Approximated Conformance Time}}$.
In this formula, a higher $ PI $ value means conformance is computed in less time.
As all our proposed methods need a preprocessing phase (e.g., for clustering the traces), we compute the $ PI$ with and without the preprocessing phase.

The accuracy of the approximation, i.e., the difference between approximated conformance value and the actual fitness value shows how close is the approximated fitness to the actual fitness value that is computed by $ Accuracy= |AccFitness - AppxFitness|$.  
Also, we measure the distance of the provided upper and lower bounds.
The bound width of an approximation is computed by $BoundWidth= UBFitness - LBFitness $. 
Tighter bound widthes means that we have more accurate bounds. 

\vspace{-0.1cm}
\subsection{Experimental Result and Discussion}
In \autoref{fig:PerfImprove}, we show how different approximation methods improve the performance of conformance checking. 
For most of the cases, the improvement is higher for the \textit{simulation} method.
This is because, the most time consuming part in conformance checking is computing the optimal alignment.
As in the \textit{simulation} method, there is no need to do any alignment computation, it is faster than other methods.
For some event logs, the \textit{statistical} sampling method~\cite{bauere_2019_stimating} is not able to provide the approximation faster than the normal conformance checking (i.e., $PI < 1 $). 
It happens because, this method is not able to benefit from the parallel computing of alignment and after each alignment computation it needs to check if it needs to do more alignment or not.  
For the \textit{statistical} method, decreasing approximation parameter leads to more precise approximations; however, it causes to have less improvement. 
Among the \textit{candidate selection} methods, using the \textit{frequency} method usually leads to a higher performance improvement. 

For some event logs, e.g., \textit{Road}, none of the method has a high \textit{PI} value. 
It happens because in \autoref{fig:PerfImprove}, we consider the preprocessing time.
The preprocessing time corresponds to choosing the candidate traces and simulating the process model behaviors that needs to be done ones per each event log or process model.
For the candidate selection methods, this phase is independent of process models and for doing that we do not need to consider any process model.
For the simulation method, this phase is independent of the given event log.
Thus, we are able to do the preprocessing step before conformance approximation. 
If we use some event log standards such as MXML and Parquet, we do not need to preprocess the event log for the \textit{frequency} and \textit{random} method because we know the number of variants and their frequency beforehand.

In \autoref{fig:PerfImprove2}, we show the performance improvement without considering the preprocessing time.
As the \textit{statistical} sampling method does not have preprocessing phase, it is not shown in this figure.
It is shown that there is a linear decrement in improvement of the \textit{candidate selection} methods by increasing the approximation parameter. 
It is expectable, as increasing in this parameter for candidate selection methods means more optimal alignment computations that requires more time. 
For example, by considering $ 5 $ for this parameter, means that we need to compute $ 5\% $ of all optimal alignments of the normal conformance checking.
Therefore, it is expected that the approximated conformance value will be computed in $ 20 $ times faster than using normal alignment.

\begin{table}[tb]
	
	\centering
	\caption{The average of approximation accuracy for conformance values when we use different approximation methods. Here we used different Inductive miner thresholds.
	 }
	\label{tab:Acc}
	\vspace{-0.1cm}
	\includegraphics[width=0.66\textwidth] {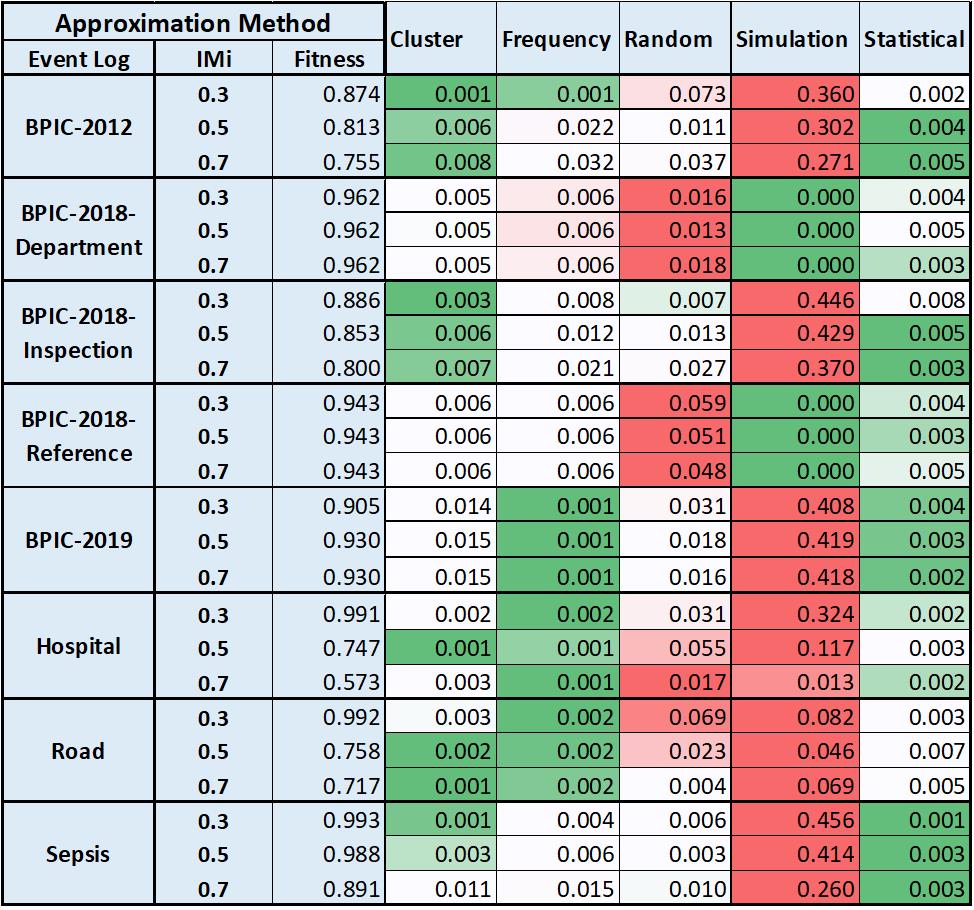}
	\vspace{-0.1cm}

\end{table}

After analyzing the performance improvement capabilities of the proposed methods, in \autoref{tab:Acc}, we compare the accuracy of their approximations. 
In this regards, the average accuracy values of the approximated conformance values are shown in this table.
The lower value means a higher accuracy or in other words, the approximated fitness value is closer to the actual fitness value. 
In this table, \textit{AFit} shows the actual fitness value when the normal conformance checking method is used.
We used different values for the approximation parameter as explained in \autoref{subsec:ES}.
The results show that for most of the event logs the accuracy of the \textit{simulation} method is not good enough.
However, for \textit{BPIC-}$ 2018 $\textit{-Reference} and \textit{BPIC-}$ 2018 $\textit{-Department}, that have simpler process models, using this method, we generated almost all the model behavior (i.e., $ \phi_v $) and obtain perfect accuracy.
Results show that if we use the \textit{statistical}, and \textit{frequency} methods, we usually obtain accuracy value below $ 0.01 $ which is acceptable for many applications.
Among the above methods, results of the statistical sampling method are more stable accuracy and returns accurate results.
However, the accuracy of candidate selection methods is usually improved by using a higher approximation parameter.

\begin{figure*}[tb]
	\begin{center}		
		\includegraphics[width=0.95\textwidth,height=4.8cm ]{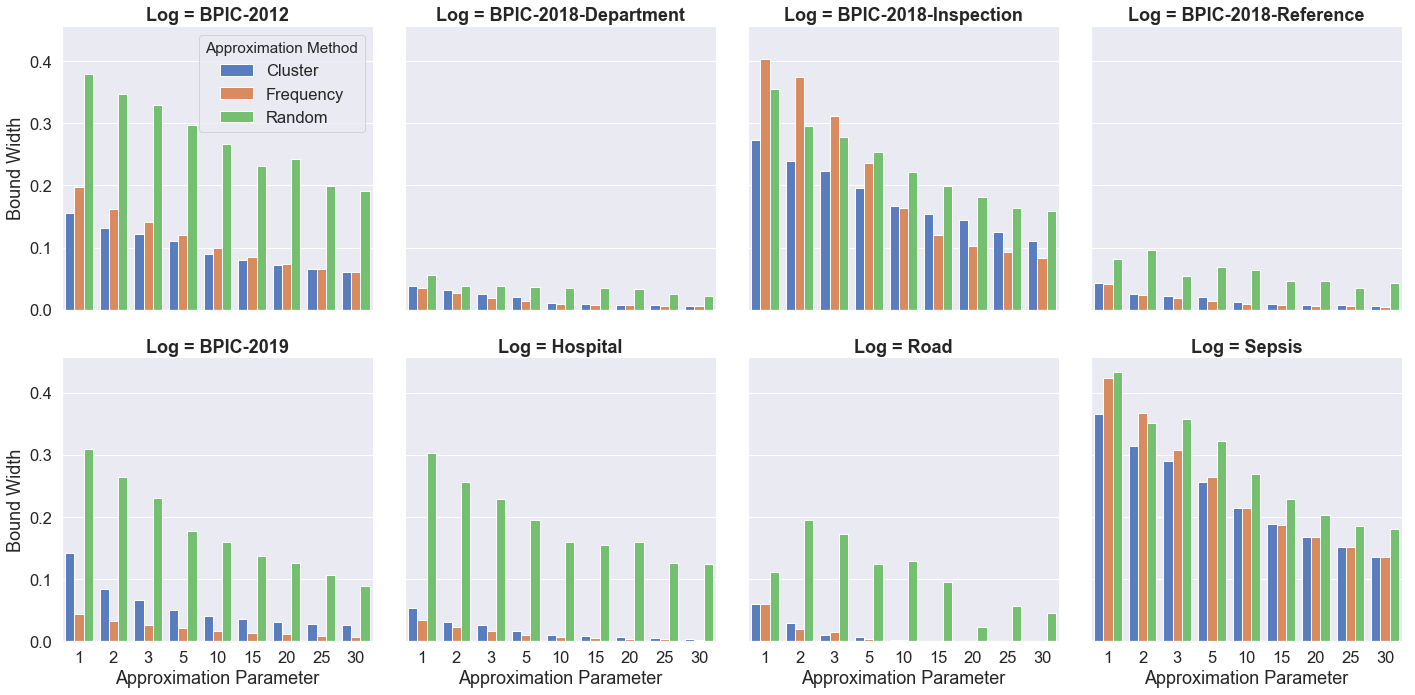}
		\caption{The average of bound width using different approximation methods.
			\label{fig:BoundWidth}
		}
	\end{center}
		\vspace{-0.3cm}
\end{figure*}

In the next experiment, we aim to evaluate the provided bounds for the approximation.
\autoref{fig:BoundWidth} shows how increasing the value of approximation parameter increases the accuracy of the provided lower and upper bounds. 
As the \textit{statistical} method does not provide any bounds, we do not consider it in this experiment. 
The \textit{simulation} method is not able to provide tight bound widths for most of the event logs.
For most of the event logs, the \textit{frequency} method results in tighter bounds. 
However, for event logs like \textit{Sepsis} for which there is no high frequent trace-variant, the \textit{clustering} method provides more accurate bounds. 
If there are high frequent variants in the event log, it is recommended to use the \textit{frequency} approximation method.
Note that, for all methods, by increasing the value of approximation parameter, we decrease the bound width.

Considering both \autoref{fig:PerfImprove2} and \autoref{fig:BoundWidth}, we observe that there is a trade off between the performance and the accuracy of the approximation methods. 
By increasing the number of visible traces in $ M_B $, we need more time to approximate the fitness value; but, we will provide more accurate bounds. 
In the case that we set the approximation parameter to $ 100 $, the bound width will be zero; however, there will not any improvement in performance of the conformance checking. 
by adjusting the approximation parameter, the end user is able to specify the performance improvement.

\autoref{fig:BoundWidth} shows that for some event logs like \textit{Sepsis} and $ \textit{BPIC-}2018\textit{-Inspection}$, none of the approximation methods are able to provide tight bounds. 
That happens because in these event logs not only do we have lots of unique traces; but, also these traces are not close to each other. 
In \autoref{tab:Similarity}, we show the average of edit distance of the most similar trace in the event logs that equals to $\displaystyle{Average_{\sigma \in L}\space\Phi(\sigma, L-\sigma)} $.
If the traces in an event log are similar to each other, we are able to provide tight bounds by the approximation methods.
This characteristic of the event log can be analyzed without any process model before the approximation. 
Therefore, it is expected to use more traces in $ M_B $ when the traces are not similar.
Using this preprocessing step, user is able to adjust the approximation parameter easier.

\begin{table*}[tb]
	\caption{The average similarity of traces in different event logs.}
	\vspace{-0.1cm}
	\label{tab:Similarity}
	\center
	\scriptsize
		\vspace{-0.2cm}
	\begin{tabular}{|l|l|l|l|l|l|l|l|}
		\hline
	
		BPIC-2012 & Department & Inspection & References & BPIC-2019 & Hospital & Road & Sepsis \\ \hline
		3.686 
		& 1.224
		& 3.269
		& 1.481
		& 5.108
		& 1.745
		& 1.113
		& 3.956
		\\ \hline
	\end{tabular}
	\vspace{-0.2cm}
\end{table*}

Finally, we analyze the accuracy of the provided information about deviations. 
We first analyze the normal alignments of event logs and process models. 
Thereafter for each alignment, we determine the six most problematic activities based on their deviation ratio that is computed based on the following formula.

\scriptsize
\begin{equation}
DeviationRatio=\frac{\textit{AsynchronousMoves}}{\textit{AsynchronousMoves} + \textit{SynchronousMoves}}
\end{equation}
\normalsize

Afterward, we compared the deviation ratio of these problematic activities with the case that the approximation method was used.
The result of this experiment is given in \autoref{tab: Deviation}.
In this experiment, we used the \textit{frequency} method with the approximation parameter equals to $10$.
We did not compare the result with the \textit{statistical} method as the goal of this method is either the fitness value or the number of asynchronous moves; but, could not return both of them at the same time\footnote{Approximating deviations is required much more time using the \textit{statistical} method. }. 
Results show that using the \textit{frequency} method, we find the problematic activities that have high deviation rates.

Considering all the experiments, we conclude that using frequency of traces for selecting candidates is more practical.
Moreover,  the candidate selection methods give more flexibility to users to trade of between the performance and the accuracy of approximations compared to the \textit{statistical} method that some times could not improve the performance and has nondeterministic results. 
In addition, the proposed methods provide bounds for the approximated alignment value and deviation rates for activities that is useful for many diagnostic applications.
Finally, the proposed methods are able to use parallel computation and benefit from adjusted computational resources.

\begin{table*}[tb]
	\centering
	\center
	\caption{Comparison of deviation ratio of the six most problematic activities using normal alignment (\textit{Real}) and the \textit{frequency} based approximation method (\textit{Appx}). 
	}
\vspace{-0. cm}
	\label{tab: Deviation}
	\scriptsize
	\begin{tabular}{|l|r|r|r|r|r|r|r|r|r|r|r|r|r|r|r|r|}
		\cmidrule{2-17}    \multicolumn{1}{r|}{} & \multicolumn{2}{c|}{\cellcolor[rgb]{ .867,  .922,  .969}BPIC-2012} & \multicolumn{2}{c|}{\cellcolor[rgb]{ .867,  .922,  .969}Department} & \multicolumn{2}{c|}{\cellcolor[rgb]{ .867,  .922,  .969}Inspection} & \multicolumn{2}{c|}{\cellcolor[rgb]{ .867,  .922,  .969}References} & \multicolumn{2}{c|}{\cellcolor[rgb]{ .867,  .922,  .969}BPIC-2019} & \multicolumn{2}{c|}{\cellcolor[rgb]{ .867,  .922,  .969}Hospital} & \multicolumn{2}{c|}{\cellcolor[rgb]{ .867,  .922,  .969}Road} & \multicolumn{2}{c|}{\cellcolor[rgb]{ .867,  .922,  .969}Sepsis} \\
		\cmidrule{2-17}    \multicolumn{1}{r|}{} & \multicolumn{1}{l|}{\cellcolor[rgb]{ .886,  .937,  .855}Appx} & \multicolumn{1}{l|}{\cellcolor[rgb]{ .851,  .882,  .949}Real} & \multicolumn{1}{l|}{\cellcolor[rgb]{ .886,  .937,  .855}Appx} & \multicolumn{1}{l|}{\cellcolor[rgb]{ .851,  .882,  .949}Real} & \multicolumn{1}{l|}{\cellcolor[rgb]{ .886,  .937,  .855}Appx} & \multicolumn{1}{l|}{\cellcolor[rgb]{ .851,  .882,  .949}Real} & \multicolumn{1}{l|}{\cellcolor[rgb]{ .886,  .937,  .855}Appx} & \multicolumn{1}{l|}{\cellcolor[rgb]{ .851,  .882,  .949}Real} & \multicolumn{1}{l|}{\cellcolor[rgb]{ .886,  .937,  .855}Appx} & \multicolumn{1}{l|}{\cellcolor[rgb]{ .851,  .882,  .949}Real} & \multicolumn{1}{l|}{\cellcolor[rgb]{ .886,  .937,  .855}Appx} & \multicolumn{1}{l|}{\cellcolor[rgb]{ .851,  .882,  .949}Real} & \multicolumn{1}{l|}{\cellcolor[rgb]{ .886,  .937,  .855}Appx} & \multicolumn{1}{l|}{\cellcolor[rgb]{ .851,  .882,  .949}Real} & \multicolumn{1}{l|}{\cellcolor[rgb]{ .886,  .937,  .855}Appx} & \multicolumn{1}{l|}{\cellcolor[rgb]{ .851,  .882,  .949}Real} \\
		\midrule
		\rowcolor[rgb]{ .867,  .922,  .969} Activity 1  & \cellcolor[rgb]{ .886,  .937,  .855}1.00 & \cellcolor[rgb]{ .851,  .882,  .949}1.00 & \cellcolor[rgb]{ .886,  .937,  .855}1.00 & \cellcolor[rgb]{ .851,  .882,  .949}1.00 & \cellcolor[rgb]{ .886,  .937,  .855}1.00 & \cellcolor[rgb]{ .851,  .882,  .949}1.00 & \cellcolor[rgb]{ .886,  .937,  .855}1.00 & \cellcolor[rgb]{ .851,  .882,  .949}1.00 & \cellcolor[rgb]{ .886,  .937,  .855}1.00 & \cellcolor[rgb]{ .851,  .882,  .949}1.00 & \cellcolor[rgb]{ .886,  .937,  .855}1.00 & \cellcolor[rgb]{ .851,  .882,  .949}0.99 & \cellcolor[rgb]{ .886,  .937,  .855}1.00 & \cellcolor[rgb]{ .851,  .882,  .949}0.96 & \cellcolor[rgb]{ .886,  .937,  .855}1.00 & \cellcolor[rgb]{ .851,  .882,  .949}1.00 \\
		\cmidrule{2-17}    \rowcolor[rgb]{ .867,  .922,  .969} Activity 2 & \cellcolor[rgb]{ .886,  .937,  .855}1.00 & \cellcolor[rgb]{ .851,  .882,  .949}1.00 & \cellcolor[rgb]{ .886,  .937,  .855}0.53 & \cellcolor[rgb]{ .851,  .882,  .949}0.53 & \cellcolor[rgb]{ .886,  .937,  .855}1.00 & \cellcolor[rgb]{ .851,  .882,  .949}1.00 & \cellcolor[rgb]{ .886,  .937,  .855}1.00 & \cellcolor[rgb]{ .851,  .882,  .949}0.50 & \cellcolor[rgb]{ .886,  .937,  .855}1.00 & \cellcolor[rgb]{ .851,  .882,  .949}1.00 & \cellcolor[rgb]{ .886,  .937,  .855}1.00 & \cellcolor[rgb]{ .851,  .882,  .949}0.95 & \cellcolor[rgb]{ .886,  .937,  .855}1.00 & \cellcolor[rgb]{ .851,  .882,  .949}0.96 & \cellcolor[rgb]{ .886,  .937,  .855}1.00 & \cellcolor[rgb]{ .851,  .882,  .949}1.00 \\
		\cmidrule{2-17}    \rowcolor[rgb]{ .867,  .922,  .969} Activity 3 & \cellcolor[rgb]{ .886,  .937,  .855}1.00 & \cellcolor[rgb]{ .851,  .882,  .949}0.94 & \cellcolor[rgb]{ .886,  .937,  .855}0.37 & \cellcolor[rgb]{ .851,  .882,  .949}0.37 & \cellcolor[rgb]{ .886,  .937,  .855}1.00 & \cellcolor[rgb]{ .851,  .882,  .949}1.00 & \cellcolor[rgb]{ .886,  .937,  .855}0.31 & \cellcolor[rgb]{ .851,  .882,  .949}0.28 & \cellcolor[rgb]{ .886,  .937,  .855}1.00 & \cellcolor[rgb]{ .851,  .882,  .949}0.98 & \cellcolor[rgb]{ .886,  .937,  .855}0.96 & \cellcolor[rgb]{ .851,  .882,  .949}0.95 & \cellcolor[rgb]{ .886,  .937,  .855}1.00 & \cellcolor[rgb]{ .851,  .882,  .949}0.83 & \cellcolor[rgb]{ .886,  .937,  .855}0.59 & \cellcolor[rgb]{ .851,  .882,  .949}0.48 \\
		\cmidrule{2-17}    \rowcolor[rgb]{ .867,  .922,  .969} Activity 4 & \cellcolor[rgb]{ .886,  .937,  .855}0.64 & \cellcolor[rgb]{ .851,  .882,  .949}0.45 & \cellcolor[rgb]{ .886,  .937,  .855}0.06 & \cellcolor[rgb]{ .851,  .882,  .949}0.06 & \cellcolor[rgb]{ .886,  .937,  .855}1.00 & \cellcolor[rgb]{ .851,  .882,  .949}0.85 & \cellcolor[rgb]{ .886,  .937,  .855}0.00 & \cellcolor[rgb]{ .851,  .882,  .949}0.00 & \cellcolor[rgb]{ .886,  .937,  .855}1.00 & \cellcolor[rgb]{ .851,  .882,  .949}0.91 & \cellcolor[rgb]{ .886,  .937,  .855}1.00 & \cellcolor[rgb]{ .851,  .882,  .949}0.88 & \cellcolor[rgb]{ .886,  .937,  .855}1.00 & \cellcolor[rgb]{ .851,  .882,  .949}0.82 & \cellcolor[rgb]{ .886,  .937,  .855}0.43 & \cellcolor[rgb]{ .851,  .882,  .949}0.32 \\
		\cmidrule{2-17}    \rowcolor[rgb]{ .867,  .922,  .969} Activity 5 & \cellcolor[rgb]{ .886,  .937,  .855}0.16 & \cellcolor[rgb]{ .851,  .882,  .949}0.01 & \cellcolor[rgb]{ .886,  .937,  .855}0.00 & \cellcolor[rgb]{ .851,  .882,  .949}0.00 & \cellcolor[rgb]{ .886,  .937,  .855}0.58 & \cellcolor[rgb]{ .851,  .882,  .949}0.40 & \cellcolor[rgb]{ .886,  .937,  .855}0.00 & \cellcolor[rgb]{ .851,  .882,  .949}0.00 & \cellcolor[rgb]{ .886,  .937,  .855}0.13 & \cellcolor[rgb]{ .851,  .882,  .949}0.11 & \cellcolor[rgb]{ .886,  .937,  .855}0.83 & \cellcolor[rgb]{ .851,  .882,  .949}0.82 & \cellcolor[rgb]{ .886,  .937,  .855}1.00 & \cellcolor[rgb]{ .851,  .882,  .949}0.72 & \cellcolor[rgb]{ .886,  .937,  .855}0.20 & \cellcolor[rgb]{ .851,  .882,  .949}0.25 \\
		\cmidrule{2-17}    \rowcolor[rgb]{ .867,  .922,  .969} Activity 6 & \cellcolor[rgb]{ .886,  .937,  .855}0.67 & \cellcolor[rgb]{ .851,  .882,  .949}0.00 & \cellcolor[rgb]{ .886,  .937,  .855}1.00 & \cellcolor[rgb]{ .851,  .882,  .949}0.00 & \cellcolor[rgb]{ .886,  .937,  .855}0.29 & \cellcolor[rgb]{ .851,  .882,  .949}0.16 & \cellcolor[rgb]{ .886,  .937,  .855}0.01 & \cellcolor[rgb]{ .851,  .882,  .949}0.00 & \cellcolor[rgb]{ .886,  .937,  .855}0.13 & \cellcolor[rgb]{ .851,  .882,  .949}0.11 & \cellcolor[rgb]{ .886,  .937,  .855}0.82 & \cellcolor[rgb]{ .851,  .882,  .949}0.82 & \cellcolor[rgb]{ .886,  .937,  .855}0.10 & \cellcolor[rgb]{ .851,  .882,  .949}0.10 & \cellcolor[rgb]{ .886,  .937,  .855}0.27 & \cellcolor[rgb]{ .851,  .882,  .949}0.22 \\
		\bottomrule
	\end{tabular}%
	\vspace{-0.4cm}
	\label{tab:addlabel}%
\end{table*}%

\vspace{-0.1cm}
\section{Conclusion}
\label{sec:conclusion}

In this paper, we proposed approximation methods for conformance value including providing upper and lower bounds. 
Instead of computing the accurate alignment between the process model and all the traces available in the event log, we propose to just consider a subset of possible behavior in the process model and use it for approximating the conformance value using the edit distance function. 
We can find this subset by computing the optimal alignments of some candidate traces in the event log or by simulating the process model.
To evaluate our proposed methods, we developed them in ProM framework and also imported them to RapidProM and applied them on several real event logs.
Results show that these methods decrease the conformance checking time and at the same time find approximated values close to the actual alignment value. 
We found that the \textit{simulation} method is suitable to be used when the given process model is simple. 
We also show that using the \textit{frequency} method is more applicable to select the candidate traces and have accurate results. 
Results also indicate that although the \textit{statistical} method is able to approximate accurately, it takes more time and for some event logs, it is slower than the normal conformance checking. 

As future work, we aim to find out what the best subset selection method is due to the available time and event data.
Also, it is possible to provide an incremental approximation tool that increases the $ M_B $ during the time and let the end user decide when the accuracy is enough.
Here, we did not use the probabilities for the simulation method, we think that by using the distribution in the event log, we enhance the \textit{simulation} method. 

\vspace{-0.1cm}
	\bibliographystyle{splncs}
	\bibliography{bibliography}

\end{document}